\documentclass[conference]{IEEEtran}
\IEEEoverridecommandlockouts
\usepackage{cite}
\usepackage{amsmath,amssymb,amsfonts}
\usepackage{algorithmic}
\usepackage{graphicx}
\usepackage{textcomp}
\usepackage{xcolor}
\usepackage{booktabs} 
\def\BibTeX{{\rm B\kern-.05em{\sc i\kern-.025em b}\kern-.08em
    T\kern-.1667em\lower.7ex\hbox{E}\kern-.125emX}}
\begin{document}

\title{Model Compression Engine for Wearable Devices Skin Cancer Diagnosis\\
\thanks{NSF Grant OIA-1849243}
}

\author{\IEEEauthorblockN{1\textsuperscript{st} Jacob M. Delgado-López}
\IEEEauthorblockA{\textit{Department of Computer Science and Engineering} \\
\textit{University of Puerto Rico Mayagüez}\\
Mayagüez, Puerto Rico \\
jacob.delgado@upr.edu}
\and
\IEEEauthorblockN{2\textsuperscript{nd} Andrea P. Seda-Hernandez}
\IEEEauthorblockA{\textit{Department of Computer Science and Engineering} \\
\textit{University of Puerto Rico Mayagüez}\\
Mayagüez, Puerto Rico \\
andrea.seda3@upr.edu}
\and
\IEEEauthorblockN{3\textsuperscript{rd} Juan D. Guadalupe-Rosado}
\IEEEauthorblockA{\textit{Department of Computer Science and Engineering} \\
\textit{University of Puerto Rico Mayagüez}\\
Mayagüez, Puerto Rico \\
juan.guadalupe@upr.edu}
\and
\IEEEauthorblockN{4\textsuperscript{th} Luis E. Fernandez Ramirez}
\IEEEauthorblockA{\textit{Department of Computer Science and Engineering} \\
\textit{University of Puerto Rico Mayagüez}\\
Mayagüez, Puerto Rico \\
luis.fernandez20@upr.edu}
\and
\IEEEauthorblockN{5\textsuperscript{th} Miguel Giboyeaux-Camilo}
\IEEEauthorblockA{\textit{Department of Computer Science and Engineering} \\
\textit{University of Puerto Rico Mayagüez}\\
Mayagüez, Puerto Rico \\
miguel.giboyeaux@upr.edu}
\and
\IEEEauthorblockN{6\textsuperscript{th} Wilfredo E. Lugo-Beauchamp}
\IEEEauthorblockA{\textit{Department of Computer Science and Engineering} \\
\textit{University of Puerto Rico Mayagüez}\\
Mayagüez, Puerto Rico \\
wilfredo.lugo1@upr.edu}
}

\maketitle

\begin{abstract}
Skin cancer is one of the most prevalent and preventable types of cancer, yet its early detection remains a challenge, particularly in resource-limited settings where access to specialized healthcare is scarce. This study proposes an AI-driven diagnostic tool optimized for embedded systems to address this gap. Using transfer learning with the MobileNetV2 architecture, the model was adapted for binary classification of skin lesions into "Skin Cancer" and "Other." The TensorRT framework was employed to compress and optimize the model for deployment on the NVIDIA Jetson Orin Nano, balancing performance with energy efficiency. Comprehensive evaluations were conducted across multiple benchmarks, including model size, inference speed, throughput, and power consumption. The optimized models maintained their performance, achieving an F1-Score of 87.18\% with a precision of 93.18\% and recall of 81.91\%. Post-compression results showed reductions in model size of up to 0.41, along with improvements in inference speed and throughput, and a decrease in energy consumption of up to 0.93 in INT8 precision. These findings validate the feasibility of deploying high-performing, energy-efficient diagnostic tools on resource-constrained edge devices. Beyond skin cancer detection, the methodologies applied in this research have broader applications in other medical diagnostics and domains requiring accessible, efficient AI solutions. This study underscores the potential of optimized AI systems to revolutionize healthcare diagnostics, thereby bridging the divide between advanced technology and underserved regions.
\end{abstract}


\section{Introduction}
Skin cancer is one of the most common and yet preventable types of cancer, affecting millions of people worldwide each year \cite{urban_global_2021}. According to recent estimates, more than 5 million new cases are diagnosed annually in the United States alone \cite{cdc_melanoma_2024}. Early detection is essential to improve patient outcomes, as the survival rate decreases dramatically once cancer metastasizes \cite{sandru_survival_2014} The survival rate decreases even more in underserved areas due to the lack of necessary resources and expertise to provide timely and accurate diagnoses.

Diagnostic challenges in skin cancer are compounded by the fact that visual inspection by dermatologists, while effective, is often limited by subjectivity and the availability of expert clinicians. Advanced diagnostic methods, such as biopsy and histopathological analysis, although more accurate, are invasive and time-consuming \cite{noauthor_skin_nodate}. This diagnostic gap is particularly pronounced in regions with limited access to dermatological care or advanced diagnostic equipment. In such settings, delayed diagnosis can significantly impact patient outcomes, highlighting the need for portable, efficient, and accessible diagnostic tools that can be deployed in resource-limited environments.

By leveraging Artificial intelligence (AI), it is possible to reduce the complications in the traditional methods while improving the diagnosis rate \cite{mahmoud_early_2024}, \cite{beltrami_artificial_2022}.  Through optimization, AI models can be adapted for deployment on edge devices, balancing performance with energy efficiency. This approach enables real-time diagnosis, bringing sophisticated diagnostic capabilities to remote areas where traditional healthcare resources are not available.

This work aims to address the diagnostic challenges of skin cancer detection, particularly in low-resource settings, by developing AI-driven diagnostic tools optimized for embedded systems. These tools can provide rapid, accurate, and accessible diagnostics, reducing the dependency on traditional high-performance technologies with intensive resource use. Furthermore, the methods developed in this research have broader implications beyond skin cancer, with potential applications in other medical fields and sectors such as soil classification and autonomous robots.

To establish a foundation for this work, a baseline model for skin lesion detection was developed using the MobileNetV2 pre-trained deep-learning architecture. Subsequently, the model underwent compression and optimization using the TensorRT framework, enabling its deployment on an NVIDIA Jetson Orin Nano. The performance of the optimized model was rigorously evaluated across metrics such as precision, recall, model size, inference time, and power consumption. The results underscore the viability of deploying AI-powered diagnostic tools in resource-constrained environments, highlighting their potential to bridge critical gaps in global healthcare accessibility.

Previous research has demonstrated the development of highly accurate models for skin cancer diagnosis \cite{mahmoud_early_2024}, \cite{beltrami_artificial_2022}. However, much of this work remains confined to controlled environments, with limited progression toward real-world clinical implementation \cite{brancaccio_artificial_2024}. The DermaSensor is the first FDA-approved AI-powered device for skin cancer detection \cite{noauthor_homepage_nodate}, as far as the authors are aware. Unlike the research presented here, the DermaSensor relies on spectral scans rather than image-based analysis. This distinction highlights the novelty of this study, which focuses on deploying an image-based diagnostic model directly on an embedded system, paving the way for practical and accessible real-world applications.

\section{Methods}

This study investigates the feasibility of developing an AI-driven diagnostic tool for skin cancer detection that is both accurate and optimized for deployment in resource-constrained environments. The methodology encompasses dataset preparation, model training, and model optimization to achieve efficient performance on embedded systems.

\subsection{Model Training}
The backbone of the proposed diagnostic tool is MobileNetV2 \cite{howard_mobilenets_2017}, a pre-trained deep-learning architecture known for its efficiency and suitability for mobile and embedded applications. The technique of transfer learning (TL) was employed to adapt this model for the specific task of skin cancer detection. Transfer learning allows the reuse of knowledge from solving a related problem or dataset, enabling improved performance on new tasks with limited data \cite{noauthor_what_2024}. In this study, the final classification layer was retrained for 50 epochs using the Skin Lesions Classification Dataset (SLCD) \cite{noauthor_skin_nodate-1}, followed by the fine-tuning of the last four layers. 

To simplify the classification task and align it with the study’s focus, the dataset, which originally consisted of images from 14 distinct skin lesion types, was condensed into two categories: "Skin Cancer" and "Other", facilitating binary classification. Table \ref{tab:dataset_description} provides an overview of the dataset composition and distribution.

\begin{table}[htbp]
\caption{Dataset Description}
\begin{center}
\begin{tabular}{|l|c|}
\hline
\textbf{Class} & \textbf{Value} \\
\hline
\multicolumn{2}{|l|}{\textbf{Skin Cancer}} \\
\hline
\quad Basal cell carcinoma & 3323 \\
\quad Melanoma & 4522 \\
\quad Squamous cell carcinoma & 628 \\
\hline
\textbf{Skin Cancer Total} & 8473 \\
\hline
\multicolumn{2}{|l|}{\textbf{Other}} \\
\hline
\quad Actinic keratoses & 867 \\
\quad Benign keratosis-like lesions & 262 \\
\quad Chickenpox & 1125 \\
\quad Cowpox & 990 \\
\quad Dermatofibroma & 239 \\
\quad Healthy & 1710 \\
\quad HFMD & 2415 \\
\quad Measles & 825 \\
\quad Melanocytic nevi & 12875 \\
\quad Monkeypox & 4260 \\
\quad Vascular lesions & 253 \\
\hline
\textbf{Other Total} & 25821 \\
\hline
\textbf{Grand Total} & 34294 \\
\hline
\end{tabular}
\label{tab:dataset_description}
\end{center}
\end{table}

To mitigate the challenges posed by class imbalance, 5-fold cross-validation was employed \cite{noauthor_cross-validation_nodate}. This method systematically divided the dataset into five subsets, with each subset serving as the validation set once while the remaining four were used for training. This approach ensured robust evaluation by reducing the likelihood of overfitting and providing a more reliable measure of model performance. A 90-10 split was used for training and validation, respectively. The training process utilized hyperparameters fine-tuned for optimal performance, including the learning rate, batch size, and optimizer configurations, as detailed in Table 2. The training was done on an NVIDIA GeForce RTX 3090 GPU.

\begin{table}[htbp]
\caption{Hyperparameters Description}
\begin{center}
\begin{tabular}{|c|c|}
\hline
\textbf{Hyperparameters} & \textbf{Value} \\
\hline
Batch Size & 32 \\
Learning Rate & 0.001 \\
folds & 5 \\
Epochs & 50 \\
Optimizer & Adam \\
Loss & Binary Cross Entropy \\
\hline
\end{tabular}
\label{tab:Hyperparameters}
\end{center}
\end{table}

\subsection{Key Metrics}

To evaluate the effectiveness of the proposed diagnostic model, five key performance metrics were employed during the training phase: Accuracy, Precision, Recall, F1-Score, and Receiver Operating Characteristic Area Under the Curve (ROC AUC) score. These metrics were selected to provide a comprehensive assessment of the model's predictive capabilities, ensuring its reliability for real-world deployment. Each metric offers unique insights, collectively enabling a thorough evaluation of the model’s strengths and weaknesses. Here, TP, TN, FP, and FN refer to true positives, true negatives, false positives, and false negatives, respectively.

Accuracy: Accuracy represents the proportion of correctly classified instances (both TP and TN) among the total number of predictions. It provides an overall measure of the model's correctness, offering a straightforward indication of its performance across all classes.

\begin{equation}
\text{Accuracy} = \frac{\text{TP} + \text{TN}}{\text{TP} + \text{TN} + \text{FP} + \text{FN}} \tag{1}
\end{equation}

Precision: Precision quantifies the model’s ability to correctly identify positive cases out of all instances predicted as positive. High Precision indicates a low rate of FP, which is critical in diagnostic applications where incorrect identification of non-cancerous cases as cancerous could lead to unnecessary procedures.

\begin{equation}
\text{Precision} = \frac{\text{TP}}{\text{TP} + \text{FP}} \tag{2}
\end{equation}

Recall: Recall assesses the model’s ability to identify all actual positive cases. High Recall minimizes the occurrence of FN, which is essential in ensuring that cases of skin cancer are not misclassified.

\begin{equation}
\text{Recall} = \frac{\text{TP}}{\text{TP} + \text{FN}} \tag{3}
\end{equation}

F1-Score: The F1-Score combines Precision and Recall into a single metric by calculating their harmonic mean. It provides a balanced measure of the model’s performance, especially useful when there is an imbalance between the number of positive and negative cases in the dataset.

\begin{equation}
\text{F1-Score} = 2 \cdot \frac{\text{Precision} \cdot \text{Recall}}{\text{Precision} + \text{Recall}} \tag{4}
\end{equation}

ROC AUC Score: The ROC AUC score evaluates the model's ability to distinguish between positive and negative classes across varying classification thresholds. A high ROC AUC score indicates that the model effectively balances Recall and the True Negative Rate making it a crucial metric for assessing overall classification performance.

\begin{equation}
\text{FPR} = \frac{\text{FP}}{\text{FP} + \text{TN}} \tag{5}
\end{equation}

By monitoring these metrics throughout the training and validation phases, the study ensured the development of a diagnostic tool capable of delivering high accuracy, clinical reliability, and robust performance across different thresholds. This comprehensive evaluation framework highlights the balance between minimizing misdiagnoses (FP and FN) and maximizing overall predictive performance, ensuring its applicability in real-world scenarios.

\subsection{Model Optimization and Deployment}
To ensure the model’s suitability for resource-limited environments, optimization techniques were applied using NVIDIA’s TensorRT framework. TensorRT is a high-performance deep-learning inference library \cite{noauthor_nvidiatensorrt_2024} designed to reduce model size, improve inference speed, and lower power consumption, making it ideal for deployment on edge devices such as the NVIDIA Jetson Orin Nano \cite{noauthor_jetson_nodate}.

The Jetson was selected for its balance of computational power and energy efficiency, making it well-suited for embedded applications in low-resource settings. The optimization process involved compressing the trained MobileNetV2 model and refining its inference capabilities to meet the hardware constraints of the Jetson. The final model was evaluated on several key metrics, including inference time, model size, and power consumption.

\section{Results and Discussion}
The proposed methodology was evaluated through critical performance metrics at both pre-compression and post-compression stages, enabling a comparative analysis of model size, inference speed, throughput, and power consumption. These metrics are crucial for understanding the impact of optimization on the feasibility of deploying the model in resource-limited environments.

\subsection{Before Compression}

\begin{figure}[htbp]
\centerline{\includegraphics[width=0.5\textwidth,keepaspectratio]{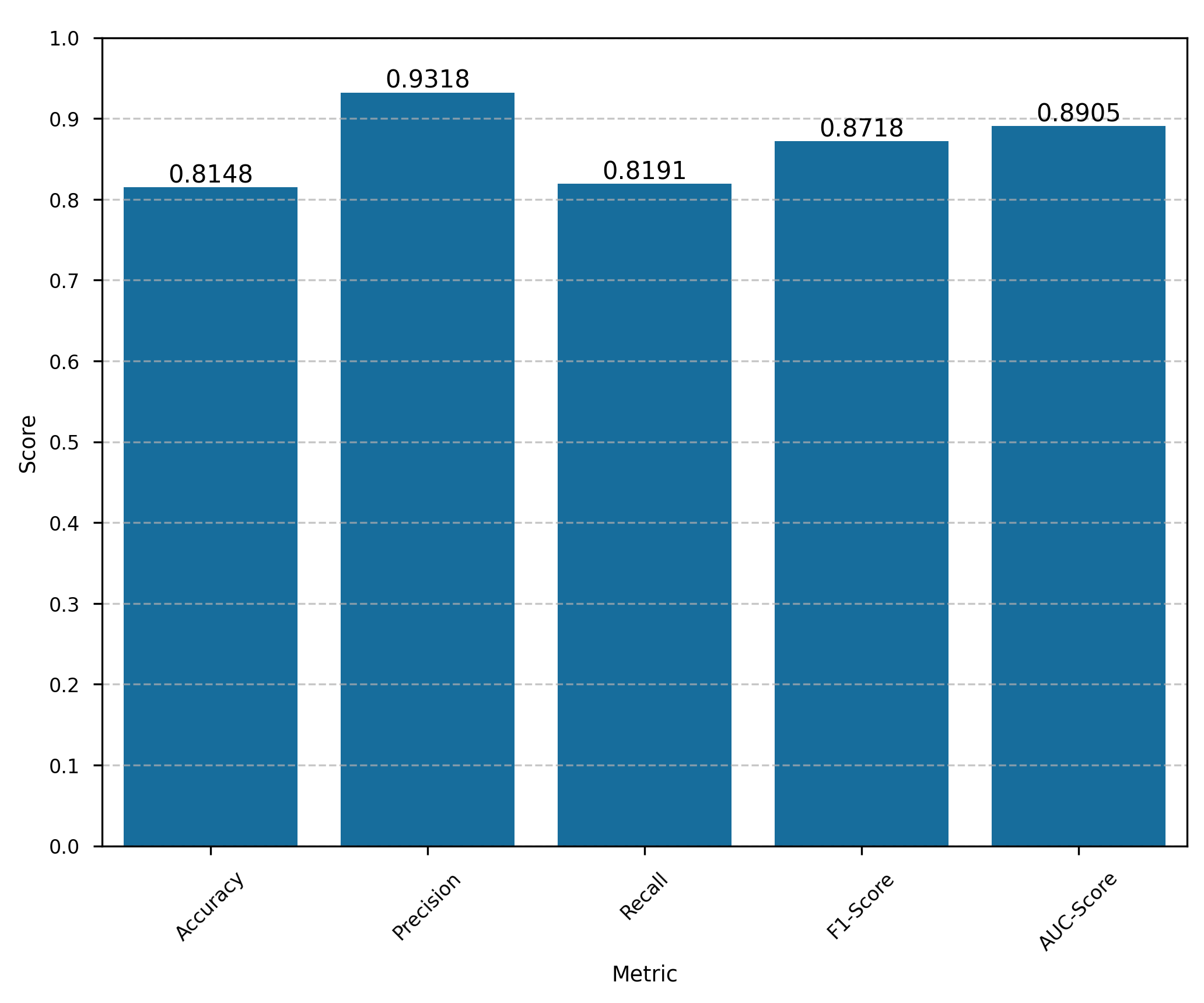}}
\caption{Performance Metrics Comparison}
\label{fig:bar_plot}
\end{figure}

The baseline performance metrics were obtained after applying TL and testing the model on a held-out dataset. These results, illustrated in Figure \ref{fig:bar_plot}, highlight the model's robustness and suitability for practical deployment scenarios requiring high accuracy and balanced performance.

The model achieved an F1-Score of 87.18\%, demonstrating its ability to effectively balance precision (93.18\%) and recall (81.91\%). This balance highlights the model's strong capability to accurately identify TP cases while minimizing FN, a critical factor in medical diagnostics where missed cases could result in delayed treatment and severe consequences for patients. Precision, at 93.18\%, indicates that the model is highly effective in avoiding FP. Meanwhile, the recall of 81.91\% shows the model’s ability to capture the majority of actual positive cases, ensuring comprehensive detection of malignant lesions. Additionally, the ROC AUC score of 89.05 underscores the model's ability to distinguish between positive and negative cases across various thresholds, reinforcing its robustness in classification performance. An overall accuracy of 81.48\% further highlights the model's consistency in delivering reliable predictions across the dataset. Together, these metrics demonstrate the model's suitability for deployment in practical diagnostic settings, where both accuracy and recall are critical for effective decision-making.

\begin{figure}[htbp]
\centerline{\includegraphics[width=0.5\textwidth,keepaspectratio]{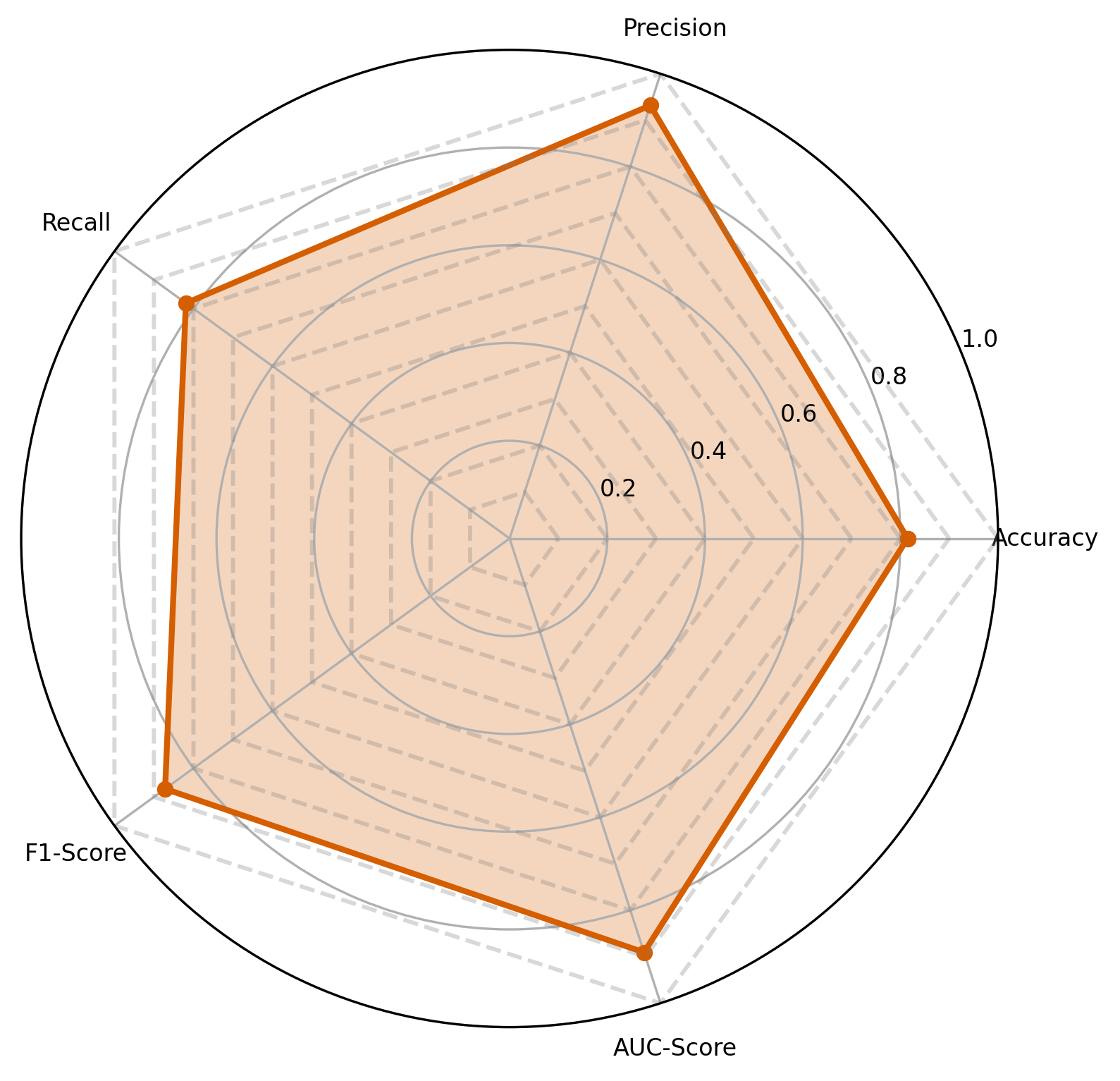}}
\caption{Performance Analysis}
\label{fig:radar_plot}
\end{figure}

Figure \ref{fig:radar_plot} further analyzes the balance among the performance metrics. While precision scored slightly higher than recall and accuracy, the metrics demonstrate overall harmony, with no significant trade-offs observed. This balance suggests that the model is well-suited for scenarios that demand robust diagnostic accuracy without sacrificing sensitivity to critical cases.

\subsection{After Compression}

The optimized models were evaluated on three primary benchmarks: model size, average inference speed, and throughput. These tests were performed on a dataset of 3,666 images using a batch size of 32, providing a robust framework for assessing the impact of the optimization. By focusing on these benchmarks, the evaluation captured critical aspects of the model’s ability to balance performance and efficiency.

\begin{figure}[htbp]
\centerline{\includegraphics[width=0.5\textwidth,keepaspectratio]{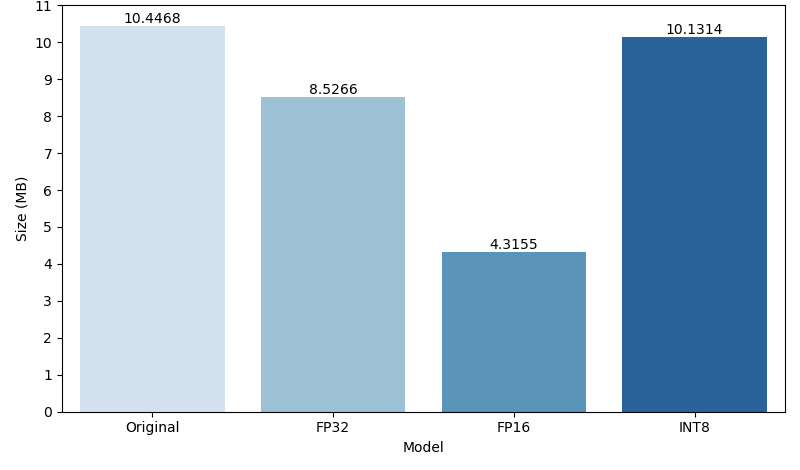}}
\caption{Post-Compression Model Size}
\label{fig:model_size}
\end{figure}

As shown in Figure \ref{fig:model_size}, TensorRT achieved significant reductions in model sizes across different precision formats. The model size decreased by factors of 0.82 for FP32, 0.41 for FP16, and 0.97 for INT8. This highlighted the framework's capability to adapt the model for edge devices with limited storage. The smaller model size translates directly to faster loading times and reduced memory overhead, improving the overall practicality of the model in real-world applications.

\begin{figure}[htbp]
\centerline{\includegraphics[width=0.5\textwidth,keepaspectratio]{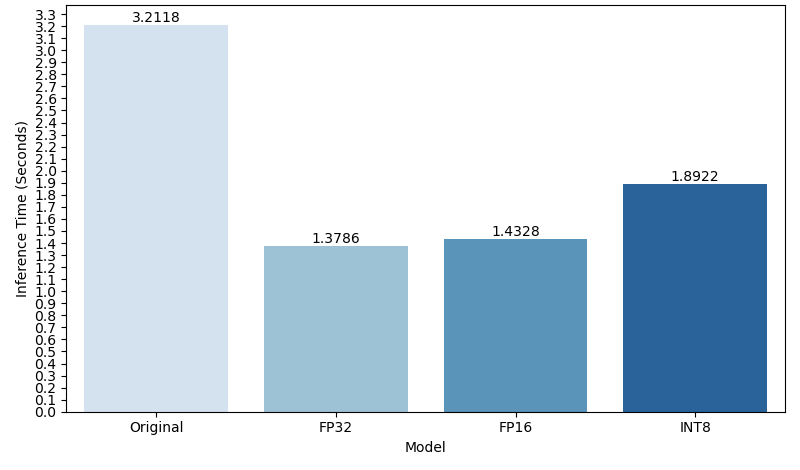}}
\caption{Post-Compression Average Inference Time}
\label{fig:model_average_inference}
\end{figure}

\begin{figure}[htbp]
\centerline{\includegraphics[width=0.5\textwidth,keepaspectratio]{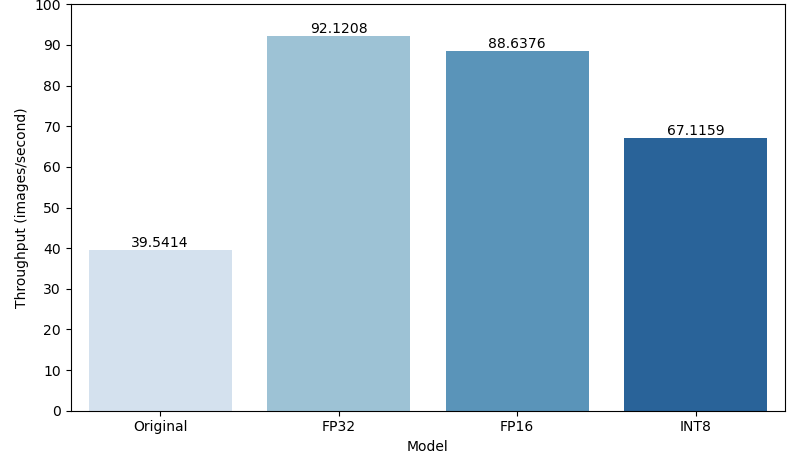}}
\caption{Post-Compression Throughput}
\label{fig:model_throughput}
\end{figure}

Figures \ref{fig:model_average_inference} and \ref{fig:model_throughput} illustrate significant enhancements in inference speed and throughput following compression. FP32 precision showed the best performance across these metrics, while FP16 underperformed slightly due to the Jetson Nano’s hardware limitations hindering the optimization process. Similarly, INT8 precision exhibited the lowest performance, a result of constrained calibration during optimization caused by the device’s 8GB RAM capacity. These limitations highlight that while TensorRT's quantization provides substantial benefits, hardware constraints play a critical role when optimizing for other precision formats.

\begin{table}[htbp]
\centering
\caption{Model Architecture Operation Distribution After Optimization}
\begin{tabular}{lrr}
\toprule
\textbf{Operation} & \multicolumn{2}{c}{\textbf{Count by Precision}} \\
\cmidrule(lr){2-3}
& FP32 & FP16/INT8 \\
\midrule
\multicolumn{3}{l}{\textit{Core Operations}} \\
Conv2D & 35 & 35 \\
DepthwiseConv2dNative & 17 & 17 \\
MatMul & 1 & 1 \\
\midrule
\multicolumn{3}{l}{\textit{Activation \& Processing}} \\
Relu6 & 35 & 35 \\
Mean & 1 & 1 \\
Mul & 17 & 18 \\
\midrule
\multicolumn{3}{l}{\textit{Utility Operations}} \\
AddV2 & 63 & 63 \\
Const & 125 & 126 \\
Pad & 4 & 4 \\
Cast\textsuperscript{*} & -- & 4 \\
\bottomrule
\multicolumn{3}{l}{\textsuperscript{*}Cast operations added for precision conversion}
\end{tabular}
\label{tab:model_architectures}
\end{table}

Table \ref{tab:model_architectures} provides an overview of the architectural modifications introduced by TensorRT during optimization. The use of mixed-precision techniques ensured that FP32 precision was retained for input and output layers, while hidden layers were optimized to FP16 or INT8 for reduced computational demand. TensorRT further streamlined operations such as Depthwise Convolution and constant tensor manipulations, minimizing computational complexity without adversely affecting accuracy. Additionally, Cast operations were introduced in FP16 and INT8 models to manage precision conversion dynamically, ensuring compatibility between layers while maintaining operational efficiency. These architectural changes collectively enhanced the model’s suitability for deployment on energy-efficient hardware.

\subsection{Power Usage}

Power usage was a critical consideration for evaluating the feasibility of deploying the optimized model in resource-limited environments, where energy efficiency is as important as computational performance. Table \ref{tab:power_analysis} provides a detailed summary of the Jetson Orin Nano's power usage across different model configurations. This evaluation included measurements during idle states and during inference, offering a comprehensive view of the energy demands associated with each precision format. 

\begin{table}[htbp]
\centering
\caption{Power Consumption Analysis of Different Model Types}
\begin{tabular}{lcr}
\toprule
\textbf{Model Type} & \textbf{Power Draw (W)} & \textbf{$\Delta$ from Idle} \\
\midrule
Original Model & 6.8 & +1.0 \\
FP32 & 6.6 & +0.8 \\
FP16 & 6.4 & +0.6 \\
INT8 & 6.3 & +0.5 \\
\midrule[\heavyrulewidth]
Idle System & 5.8 & -- \\
\bottomrule
\end{tabular}
\label{tab:power_analysis}
\end{table}

The results show that optimized models consistently consumed less power than the original model, achieving reductions of 0.97 for FP32, 0.94 for FP16, and 0.93 for INT8 formats. These improvements validate the effectiveness of TensorRT's quantization and optimization techniques in minimizing power usage while maintaining robust performance. The reduced energy consumption is particularly significant for INT8, despite its relatively lower computational efficiency, highlighting the role of mixed-precision optimization in reducing energy usage.

\section{Conclusion}
This study demonstrated the feasibility and effectiveness of deploying an AI-driven diagnostic tool for skin cancer detection on resource-constrained embedded systems. By leveraging transfer learning with the MobileNetV2 architecture and applying optimization techniques using TensorRT, the model was successfully adapted for deployment on the NVIDIA Jetson Orin Nano. The results show that the optimized models maintained an 87.18\% F1-Score, while significantly reducing model size, improving inference speed, and enhancing energy efficiency. However, some limitations were noted, particularly with FP16 and INT8 precision due to hardware constraints. Despite this, the results obtained underscored the potential of AI in addressing diagnostic gaps in resource-limited environments, where access to advanced medical infrastructure is often scarce.

Beyond skin cancer detection, the methodologies presented in this study have broader implications. The use of model compression and optimization can extend to various applications, including other medical diagnostic fields, environmental monitoring, and autonomous systems. By enabling sophisticated AI systems to operate effectively on embedded devices, this research provides a pathway toward increasing accessibility to technology in underserved regions. Future work will focus on introducing new compression techniques, such as knowledge distillation, and expanding the use case to other platforms, such as the Arduino Nano. The ultimate goal of this research is to expand the capabilities of model compression for wearable devices, enabling seamless, non-invasive diagnosis of skin cancer.

\section*{Acknowledgment}

This work would not have been possible without the support of the Edge Computing Group and the University of Puerto Rico, Mayagüez. Thank you to the NSF Grant OIA-1849243 for partially funding this research.

\nocite{*}  
\bibliographystyle{IEEEtran}
\bibliography{references}

\begin{thebibliography}{10}
\providecommand{\url}[1]{#1}
\csname url@samestyle\endcsname
\providecommand{\newblock}{\relax}
\providecommand{\bibinfo}[2]{#2}
\providecommand{\BIBentrySTDinterwordspacing}{\spaceskip=0pt\relax}
\providecommand{\BIBentryALTinterwordstretchfactor}{4}
\providecommand{\BIBentryALTinterwordspacing}{\spaceskip=\fontdimen2\font plus
\BIBentryALTinterwordstretchfactor\fontdimen3\font minus \fontdimen4\font\relax}
\providecommand{\BIBforeignlanguage}[2]{{%
\expandafter\ifx\csname l@#1\endcsname\relax
\typeout{** WARNING: IEEEtran.bst: No hyphenation pattern has been}%
\typeout{** loaded for the language `#1'. Using the pattern for}%
\typeout{** the default language instead.}%
\else
\language=\csname l@#1\endcsname
\fi
#2}}
\providecommand{\BIBdecl}{\relax}
\BIBdecl

\bibitem{urban_global_2021}
\BIBentryALTinterwordspacing
K.~Urban, S.~Mehrmal, P.~Uppal, R.~L. Giesey, and G.~R. Delost, ``The global burden of skin cancer: {A} longitudinal analysis from the {Global} {Burden} of {Disease} {Study}, 1990–2017,'' \emph{JAAD International}, vol.~2, pp. 98--108, Jan. 2021. [Online]. Available: \url{https://www.ncbi.nlm.nih.gov/pmc/articles/PMC8362234/}
\BIBentrySTDinterwordspacing

\bibitem{cdc_melanoma_2024}
\BIBentryALTinterwordspacing
CDC, ``\BIBforeignlanguage{en-us}{Melanoma of the {Skin} {Statistics}},'' Jul. 2024. [Online]. Available: \url{https://www.cdc.gov/skin-cancer/statistics/index.html}
\BIBentrySTDinterwordspacing

\bibitem{sandru_survival_2014}
\BIBentryALTinterwordspacing
A.~Sandru, S.~Voinea, E.~Panaitescu, and A.~Blidaru, ``Survival rates of patients with metastatic malignant melanoma,'' \emph{Journal of Medicine and Life}, vol.~7, no.~4, pp. 572--576, 2014. [Online]. Available: \url{https://www.ncbi.nlm.nih.gov/pmc/articles/PMC4316142/}
\BIBentrySTDinterwordspacing

\bibitem{noauthor_skin_nodate}
\BIBentryALTinterwordspacing
``Skin cancer - {Diagnosis} and treatment - {Mayo} {Clinic}.'' [Online]. Available: \url{https://www.mayoclinic.org/diseases-conditions/skin-cancer/diagnosis-treatment/drc-20377608}
\BIBentrySTDinterwordspacing

\bibitem{mahmoud_early_2024}
\BIBentryALTinterwordspacing
N.~M. Mahmoud and A.~M. Soliman, ``\BIBforeignlanguage{en}{Early automated detection system for skin cancer diagnosis using artificial intelligent techniques},'' \emph{\BIBforeignlanguage{en}{Scientific Reports}}, vol.~14, no.~1, p. 9749, Apr. 2024, publisher: Nature Publishing Group. [Online]. Available: \url{https://www.nature.com/articles/s41598-024-59783-0}
\BIBentrySTDinterwordspacing

\bibitem{beltrami_artificial_2022}
E.~J. Beltrami, A.~C. Brown, P.~J.~M. Salmon, D.~J. Leffell, J.~M. Ko, and J.~M. Grant-Kels, ``\BIBforeignlanguage{eng}{Artificial intelligence in the detection of skin cancer},'' \emph{\BIBforeignlanguage{eng}{Journal of the American Academy of Dermatology}}, vol.~87, no.~6, pp. 1336--1342, Dec. 2022.

\bibitem{brancaccio_artificial_2024}
\BIBentryALTinterwordspacing
G.~Brancaccio, A.~Balato, J.~Malvehy, S.~Puig, G.~Argenziano, and H.~Kittler, ``Artificial {Intelligence} in {Skin} {Cancer} {Diagnosis}: {A} {Reality} {Check},'' \emph{Journal of Investigative Dermatology}, vol. 144, no.~3, pp. 492--499, Mar. 2024. [Online]. Available: \url{https://www.sciencedirect.com/science/article/pii/S0022202X23029640}
\BIBentrySTDinterwordspacing

\bibitem{noauthor_homepage_nodate}
\BIBentryALTinterwordspacing
``\BIBforeignlanguage{en-US}{Homepage}.'' [Online]. Available: \url{https://www.dermasensor.com/}
\BIBentrySTDinterwordspacing

\bibitem{howard_mobilenets_2017}
\BIBentryALTinterwordspacing
A.~G. Howard, M.~Zhu, B.~Chen, D.~Kalenichenko, W.~Wang, T.~Weyand, M.~Andreetto, and H.~Adam, ``{MobileNets}: {Efficient} {Convolutional} {Neural} {Networks} for {Mobile} {Vision} {Applications},'' Apr. 2017, arXiv:1704.04861 [cs]. [Online]. Available: \url{http://arxiv.org/abs/1704.04861}
\BIBentrySTDinterwordspacing

\bibitem{noauthor_what_2024}
\BIBentryALTinterwordspacing
``\BIBforeignlanguage{en}{What is transfer learning? {\textbar} {IBM}},'' Jan. 2024. [Online]. Available: \url{https://www.ibm.com/topics/transfer-learning}
\BIBentrySTDinterwordspacing

\bibitem{noauthor_skin_nodate-1}
\BIBentryALTinterwordspacing
``\BIBforeignlanguage{en}{Skin {Lesions} {Classification} {Dataset}}.'' [Online]. Available: \url{https://www.kaggle.com/datasets/ahmedxc4/skin-ds}
\BIBentrySTDinterwordspacing

\bibitem{noauthor_cross-validation_nodate}
\BIBentryALTinterwordspacing
``Cross-{Validation} - an overview {\textbar} {ScienceDirect} {Topics}.'' [Online]. Available: \url{https://www.sciencedirect.com/topics/medicine-and-dentistry/cross-validation}
\BIBentrySTDinterwordspacing

\bibitem{noauthor_nvidiatensorrt_2024}
\BIBentryALTinterwordspacing
``{NVIDIA}/{TensorRT},'' Dec. 2024, original-date: 2019-05-02T22:02:08Z. [Online]. Available: \url{https://github.com/NVIDIA/TensorRT}
\BIBentrySTDinterwordspacing

\bibitem{noauthor_jetson_nodate}
\BIBentryALTinterwordspacing
``\BIBforeignlanguage{en-US}{Jetson {Orin} {Nano} {Developer} {Kit} {User} {Guide}}.'' [Online]. Available: \url{https://developer.nvidia.com/embedded/learn/jetson-orin-nano-devkit-user-guide/index.html}
\BIBentrySTDinterwordspacing

\end{thebibliography}

\end{document}